# Reduced Ordered Binary Decision Diagram with Implied Literals: A New knowledge Compilation Approach

**Yong Lai, Dayou Liu, Shengsheng Wang**
College of Computer Science and Technology
Jilin University, Changchun 130012, P.R. China.
e-mail: laiy07@mails.jlu.edu.cn

**Abstract**: Knowledge compilation is an approach to tackle the computational intractability of general reasoning problems. According to this approach, knowledge bases are converted off-line into a target compilation language which is tractable for on-line querying. Reduced ordered binary decision diagram (ROBDD) is one of the most influential target languages. We generalize ROBDD by associating some implied literals in each node and the new language is called reduced ordered binary decision diagram with implied literals (ROBDD-$L$). Then we discuss a kind of subsets of ROBDD-$L$ called ROBDD-$i$ with precisely $i$ implied literals ($0 \le i \le \infty$). In particular, ROBDD-0 is isomorphic to ROBDD; ROBDD-$\infty$ requires that each node should be associated by the implied literals as many as possible. We show that ROBDD-$i$ has uniqueness over some specific variables order, and ROBDD-$\infty$ is the most succinct subset in ROBDD-$L$ and can meet most of the querying requirements involved in the knowledge compilation map. Finally, we propose an ROBDD-$i$ compilation algorithm for any $i$ and a ROBDD-$\infty$ compilation algorithm. Based on them, we implement a ROBDD-$L$ package called BDDjLu and then get some conclusions from preliminary experimental results: ROBDD-$\infty$ is obviously smaller than ROBDD for all benchmarks; ROBDD-$\infty$ is smaller than the d-DNNF the benchmarks whose compilation results are relatively small; it seems that it is better to transform ROBDDs-$\infty$ into FBDDs and ROBDDs rather than straight compile the benchmarks.

## 1 Introduction

Reasoning problems in their general form are intractable and knowledge compilation has been emerging as a key direction of research for dealing with such kind of intractability [1-5]. The basic idea of knowledge compilation is to split the reasoning process into two phases: an off-line compilation phase, in which the propositional theory is compiled into some tractable target language, and an on-line query-answering phase, in which the compiled target is used to efficiently answer the queries. And the compiling time in off-line phase can be amortized by a (potentially) exponential number of on-line queries.

The target language is one of key aspects for any compilation approach. There have existed dozens of target languages so far, such as Horn theories [1], prime implicates/implicants [6, 7], reduced ordered binary decision diagram (ROBDD) [8, 9], free binary decision diagram (FBDD) [5, 10], decomposable negation normal form (DNNF, three subsets d-DNNF, DNNF$_T$ and d-DNNF$_T$) [3, 11, 12], EPCCL theory [4, 13] and so on. Therefore, it is important to choose an appropriate target language in practical applications. Darwiche and Marquis argue that the choice of a target language must be based on two key aspects: the succinctness of the target compilation language, and the class of queries and transformations that the language supports in polytime [5]. Moreover, they propose the classic knowledge compilation map, which analyzes many existing target compilation languages according to the above aspects. On this basis, some researchers extend the knowledge compilation map [12, 14, 15].

ROBDD is one of the most tractable target languages which satisfy all of querying requirements involved in the knowledge compilation map (as far as we know, the compilation languages meeting these requirements include ROBDD, MODS, d-DNNF$_T$ and EPCCL theory [5, 12, 13]) and has been quite influential in many communities such as model checking [16], AI planning [17], abductive inference [18], terminological reasoning in description logic SHIQ [19] and so on. However, ROBDD seems a bit redundant for some Boolean formulas. For example, it is well known that the ROBDD representing the Boolean formula $(x_1 \leftrightarrow y_1) \wedge \ldots \wedge (x_n \leftrightarrow y_n)$ has exponential size over the variables order $x_1 < \ldots < x_n < y_1 < \ldots < y_n$. In fact, when $x_i$ is assigned some value (*true* or *false*), $y_i$ must have the same value. This characteristic limits the application of this language to some specific areas. Therefore, in order to reduce this kind of redundancy and extend its real applications, it is very necessary to make some small changes of ROBDD without loss of too much tractability. Based on this motivation, we do the following work in this paper:

1. We add some literals called implied literals in the nodes of ROBDD meaning that the formula represented by the node implies them. We call this new target language ROBDD-$L$. Given a number $i$, we discuss a subset of ROBDD-$L$ called ROBDD-$i$ – precisely $i$ implied literals in each node. It is obvious that ROBDD-0 is isomorphic to ROBDD. Then we show that there is exactly one ROBDD-$i$ representing a given formula over a specific variables order.

2. We show that ROBDD-$\infty$ is an interesting subset of ROBDD-$L$: it is the most succinct subset of ROBDD-$L$ and we propose an algorithm which can transform every sentence of ROBDD-$L$ into an equivalent sentence of ROBDD-$\infty$ in polytime. Furthermore, we prove that ROBDD-$\infty$ can meet all the query requirements except SE mentioned in the knowledge compilation map (it is unknown whether ROBDD-$\infty$ satisfies SE or not).

3. We propose a compilation algorithm (called Build) which can compile any Boolean formula into ROBDD-$i$ ($0 \le i \le \infty$). And we optimize Build to propose a ROBDD-$\infty$ compilation algorithm called Build-inf and discuss three optimization techniques. In addition, we propose two algorithms called Inf2FBDD and Inf2ROBDD which can transform any ROBDD-∞ into FBDD and ROBDD, respectively. Combining Build, Build-inf, Inf2FBDD and Inf2ROBDD and all the operations supported by ROBDD-*L* in polytime, we devise a ROBDD-*L* package called BDDjLu and report some experimental results.

## 2 Reduced Ordered Binary Decision Diagram with Implied Literals

In the sequel X = $\{x_1, \dots, x_n\}$ is the set of Boolean variables. A Boolean formula, hereafter simply called a formula, is constructed from *true*, *false* and variables using the negation operator ¬, conjunction operator ∧ and disjunction operator ∨. A *literal* is either a variable *x* (positive literal) or its negation ¬*x* (negative literal). Given a literal *l*, its negation ¬*l* is ¬*x* if *l* is *x* and ¬*l* is *x* otherwise. A *clause C* is a set of literals representing their disjunction. *C* is a Horn clause if it contains at most one positive literal. A *term T* is a set of literals representing their conjunction. *T* is consistent iff there does not exist any variable *x* such that both *x* and ¬*x* belong to *T*. A Boolean formula in *conjunctive normal form* (CNF) is a set of clauses representing their conjunction**.** A CNF formula is Horn theory if all clauses are Horn clause. A Boolean formula in *negation normal form* (NNF) is constructed from *true*, *false* and literals using only the conjunction and disjunction operators. It is obvious that any clause, term and CNF formula is in NNF. A practical representation of NNF formula [3, 5] is a rooted, *directed acyclic graph* (DAG) where each leaf node is labeled with *true*, *false* or a literal; and each internal node is labeled with ∧ or ∨ and can have arbitrarily many children.

An *assignment A* over the variables set X (we also say that *A* is a X-assignment) is a set of literals such that *A* does not contains any literal and its negation. *A* is *complete* over X if *A* contains one and only one literal for any variable *x* in X (i.e., there exists exactly one element in $\{x, \neg x\} \cap A$ for any variable $x \in$ X), otherwise it is *partial*. It is obvious that there exists $2^{|X|}$ complete assignments over X. Any complete assignment *satisfies true* and *falsifies false*. A complete assignment *A* satisfies a literal *l* over X iff $l \in A$, *A* falsifies it otherwise; *A* satisfies a formula $\neg\varphi$ over X iff it falsifies $\varphi$, and *A* falsifies it otherwise; *A* satisfies a formula $\varphi_1 \wedge \varphi_2$ over X iff it satisfies both $\varphi_1$ and $\varphi_2$, and *A* falsifies it otherwise; *A* satisfies $\varphi_1 \vee \varphi_2$ over X iff it satisfies either $\varphi_1$ or $\varphi_2$, and *A* falsifies it otherwise. A *model M* of any formula is a complete assignment satisfies it. We call a formula *satisfiable* if it has at least one model, and we say it is *unsatisfiable* otherwise. We say a formula over X is a tautology if all complete assignments over X satisfy it. Given two formulas $\varphi_1$ and $\varphi_2$ over X, $\varphi_1$ implies $\varphi_2$ (denoted by $\varphi_1 \Rightarrow \varphi_2$) iff the models of $\varphi_1$ is subsumed by the ones of $\varphi_2$, $\varphi_1$ is equivalent to $\varphi_2$ (denoted by $\varphi_1 \Leftrightarrow \varphi_2$) iff both $\varphi_1 \Rightarrow \varphi_2$ and $\varphi_2 \Rightarrow \varphi_1$. Now we give the definition of *reduced ordered binary decision diagram with implied literals* step by step.

**Definition 1.** A *binary decision diagram with implied literals* (BDD-*L*) is a rooted DAG. Each node *v* is either *terminal* or *non-terminal* and represents some formula $\phi(v)$ in NNF. There exist two kinds of terminal nodes: *False* node (denoted by ⊥) which represents *false*, and *True* nodes labeled by a set of literals *L*(*v*) called *implied literals* which represent the term conjoining all the literals in *L*(*v*). And each non-terminal node *v* is associated with a Boolean variable *var*(*v*), implied literals *L*(*v*) and two children, called *low* child *lo*(*v*) and *high* child *hi*(*v*). Given a non-False node *v*, it is denoted by ⟨*L*(*v*)⟩ if it is a True node, otherwise it is denoted by ⟨*var*(*v*), *lo*(*v*), *hi*(*v*), *L*(*v*)⟩. At each non-terminal node *v*, *var*(*v*) does not appear in *L*(*v*), any variable appearing in *L*(*v*) does not appear in its descendent nodes, and the low (resp. high) branch is depicted as a dash (resp. solid) line corresponding to the case where the variable is assigned *false* (resp. *true*). So given a non-terminal node *v*, we have that:

$$\phi(v) = (\textstyle\bigwedge l \in L(v)) \wedge ((\neg var(v) \wedge \phi(lo(v))) \vee (var(v) \wedge \phi(hi(v)))). \qquad (*)$$

Usually, an implied literal corresponds to a simple fact implied by knowledge base, such as John does not like pink coat, Jim just like jeans and so on. In BDD-*L*, some or all simple facts are pulled out and stored explicitly at the root. From above Definition 1, we know that the only difference between BDD-*L* and BDD [8] is the implied literals. When every node in BDD-*L* is mandatory for no implied literal, then BDD-*L* is equivalent to BDD and (*) will be $\phi(v) = (\neg var(v) \wedge \phi(lo(v))) \vee (var(v) \wedge \phi(hi(v)))$. And FBDD can be seen as a special kind of BDD-0 such that each variable appears at most once on any path. For simplicity, we suppose that at least one child of a non-terminate node is non-False and the implied literals of a non-False node represent a consistent tern.

In the implementation, we use a hash table called nodes table to record all the nodes in BDD-*L*, another hash table called implied literals table, which allows us to deal with the same sets of implied literals only once in some operations (see Section 4), is used to record all the sets of implied literals in BDD-*L*, *L*(*v*) for any node *v* in the nodes table is an index pointing to some entry in implied literals table, *L*(*u*) and *L*(*v*) share the same entry if *L*(*u*) = *L*(*v*), the data unit size in implied literals table records the size of *L*(*v*) in order to facilitate model counting (see Section 4).

Nodes Table

| ID | *var* | *low* | *high* | *L* |
|---|---|---|---|---|
| 5 | $x_1$ | 3 | 4 | 0 |
| 4 | $x_2$ | 1 | 2 | 2 |
| 3 | $x_1$ | 0 | 1 | 1 |
| 2 | True | - | - | 1 |
| 1 | True | - | - | 0 |
| 0 | False | - | - | - |

Implied Literals Table

| ID | literals | size |
|---|---|---|
| 2 | $\{\neg x_1\}$ | 1 |
| 1 | $\{x_2\}$ | 1 |
| 0 | $\varnothing$ | 0 |

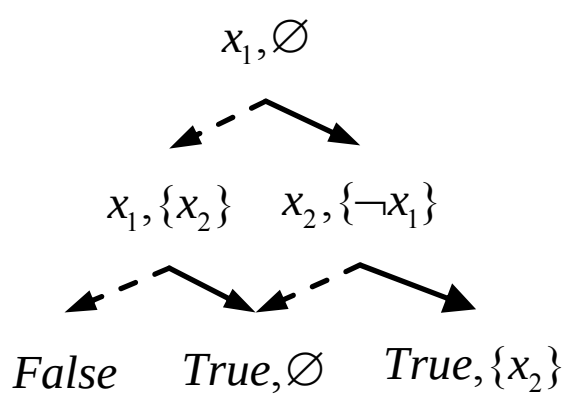

Figure 1. An example about BDD-*L*

An example about BDD-*L* is showed in Figure 1. In the following, we denote the set of the variables appearing in $v$ (i.e., $var(v)$ and the variables appearing in $L(v)$) and its descendent nodes as $VARS(v)$, the number of nodes in the BDD-*L* as $|u|$, where $u$ is the root of the BDD-*L*, $path(v)$ is a term and any literal $\neg x$ (resp. $x$) belongs to it iff there exists some node $v'$ with the variable $x$ such that its low (resp. high) branch appearing in the path from the root to $v$. The *maximal set of implied literals*, which is used frequently in this paper, is defined as follows:

**Definition 2.** Given a BDD-*L* and a non-False node $v$ in it, the maximal set of implied literals $L_\infty(v)$ is defined as follows:

$$L_\infty(v)=\begin{cases} L(v) & v \text{ is a True node;}\\ L(v)\cup\{var(v)\}\cup L_\infty(hi(v)) & lo(v)=\perp;\\ L(v)\cup\{\neg var(v)\}\cup L_\infty(lo(v)) & hi(v)=\perp;\\ L(v)\cup(L_\infty(hi(v))\cap L_\infty(lo(v))) & \text{otherwise.}\end{cases}$$

Obviously, $L(v)$ is a subset of $L_\infty(v)$ for any non-False node $v$. Given any BDD-*L*, we can compute the maximal set of implied literals for all of its nodes in polytime with the use of dynamic programming, which is used in almost all algorithms in this paper.

**Definition 3.** A BDD-*L* is *ordered* (OBDD-*L*) if

1. The set of variables is imposed over a given linear order <;
2. Given a node $u$ and its child $v$, $var(u)$ is less than any variable appearing in $v$;
3. For any non-terminal node $v$, any variable appearing in $L(v)$ is less than the ones appearing in $L_\infty(v)$ but not in $L(v)$, formally,

$$\forall x(((x\in L(v)\vee\neg x\in L(v))\wedge(x'\in L_\infty(v)\setminus L(v)\vee\neg x'\in L_\infty(v)\setminus L(v)))\rightarrow x<x').$$

From the above definition, it is obvious that BDD-*L* in Figure 1 is not ordered as it does not satisfy the condition 2 and 3 in Definition 3, and we have some simple conclusions which are used in the proofs of some propositions: given a non-False node $v$, it is easy to prove that $\phi(v)$ is satisfiable (used in the proof of Proposition 1) by induction, because at least one child of $v$ is non-False and the implied literals represent a consistent term; given any node $u$ and its child $v$ in an OBDD-*L*, $VARS(v)\subset VARS(u)$ (used in the proof of Proposition 4); given a non-False node $v$, the condition 3 is obviously satisfied if $L(u) = L_\infty(u)$ (used in the proof of Proposition 4). And the following proposition holds:

**Proposition 1.** Given any non-False node $v$ in any OBDD-*L*, each element in $L_\infty(v)$ is exactly a literal implied by $\phi(v)$, i.e., $L_\infty(v) = \{l : \phi(v) \Rightarrow l\}$.

**Proof.** By induction on the size of $|v|$. Assume that the conclusion holds for $|v| \le n$. The case $|v| = 1$ is immediately. We proceed by case analysis:

(1) $lo(v) = \perp$: We have that $L_\infty(v) = L(v) \cup \{var(v)\} \cup L_\infty(hi(v))$ from Definition 2. By induction hypothesis, $L_\infty(hi(v)) = \{l : \phi(hi(v)) \Rightarrow l\}$. This means that there exists some formula $\varphi$ such that $\phi(hi(v)) \Leftrightarrow \bigwedge(l \in L_\infty(hi(v))) \wedge \varphi$. Then by (*)

$$\phi(v)=(\bigwedge l\in L(v))\wedge(var(v)\wedge(\bigwedge l\in L_\infty(hi(v)))\wedge\varphi).$$

It is obvious that $\phi(v) \Rightarrow l$ for any $l \in L_\infty(v)$. Assume that there exists some literal $l$ such that $\phi(v) \Rightarrow l$ and $l \notin L_\infty(v)$. And any non-False node in OBDD-*L* represents a satisfiable formula. Then $\varphi \Rightarrow l$, this conflicts with the induction hypothesis.

(2) $hi(v) = \perp$: It is analogous to (1).

(3) Otherwise, we have that $L_\infty(v) = L(v) \cup (L_\infty(lo(v)) \cap L_\infty(hi(v)))$ from Definition 2. By induction hypothesis, $L_\infty(hi(v)) = \{l : \phi(hi(v)) \Rightarrow l\}$ (resp. $L_\infty(lo(v)) = \{l : \phi(lo(v)) \Rightarrow l\}$). This means that there exists some formula $\varphi$ such that $\phi(hi(v)) \Leftrightarrow \bigwedge(l \in L_\infty(hi(v))) \wedge \varphi$ (resp. $\phi(lo(v)) \Leftrightarrow \bigwedge(l \in L_\infty(lo(v))) \wedge \varphi'$). Then by (*)

$$\phi(v)=(\bigwedge l\in L(v))\wedge((\neg var(v)\wedge(\bigwedge l\in L_\infty(lo(v)))\wedge\varphi)\vee(var(v)\wedge(\bigwedge l\in L_\infty(hi(v)))\wedge\varphi')).$$

It is obvious that $\phi(v) \Rightarrow l$ for any $l \in L_\infty(v)$. Assume that there exists some literal $l$ such that $\phi(v) \Rightarrow l$ and $l \notin L_\infty(v)$. By Definition 3, then $\varphi \Rightarrow l$ or $\varphi' \Rightarrow l$, this conflicts with the induction hypothesis. ■

If we displace the OBDD-*L* with BDD-*L* in Proposition 1, then the proposition doesn't hold. The BDD-*L* in Figure 1 is a counterexample for it, where $L_\infty(v) = \varnothing$ and the formula represented by this BDD-*L* implies any literal.

**Definition 4.** An OBDD-$L$ is *reduced* (ROBDD-$L$) if

1. No two distinct nodes $u$ and $v$ have the identical variable, implied literals, low child and high child;
2. No node has two identical children;

**Definition 5.** Given a number $0 \le i \le \infty$, a OBDD-$L$ precisely has $i$ implied literals (OBDD-$i$) if (1) each node $v$ has $i$ implied literals, or (2) $j$ ($j < i$) implied literals and $L(v) = L_\infty(v)$. A ROBDD-$i$ is a reduced OBDD-$i$.

Particularly, OBDD-$\infty$ is also called "OBDD with as many implied literals as possible". It is obvious that ROBDD-0 is isomorphic to ROBDD. By Proposition 1, we have that $L(v) = L_\infty(v)$ for any node $v$ in OBDD-$\infty$, then the False node does not appear in any OBDD-$\infty$.

Figure 2 are a ROBDD-0 and a ROBDD-1 about the formula $(x_1 \leftrightarrow y_1) \wedge (x_2 \leftrightarrow y_2)$ over the variables order $x_1 < x_2 < y_1 < y_2$. The nodes in ROBDD-0 are obviously less than those in ROBDD-1. Furthermore, all the ROBDDs-$i$ ($i > 0$) presenting this formula over the same variables order are the same. If this formula is extended to $(x_1 \leftrightarrow y_1) \wedge \ldots \wedge (x_n \leftrightarrow y_n)$ over $x_1 < \ldots < x_n < y_1 < \ldots < y_n$, then ROBDD-0 will have more than $2^{n+1}$ nodes, while the number of nodes in ROBDD-$i$ ($i > 0$) is $2n + 1$.

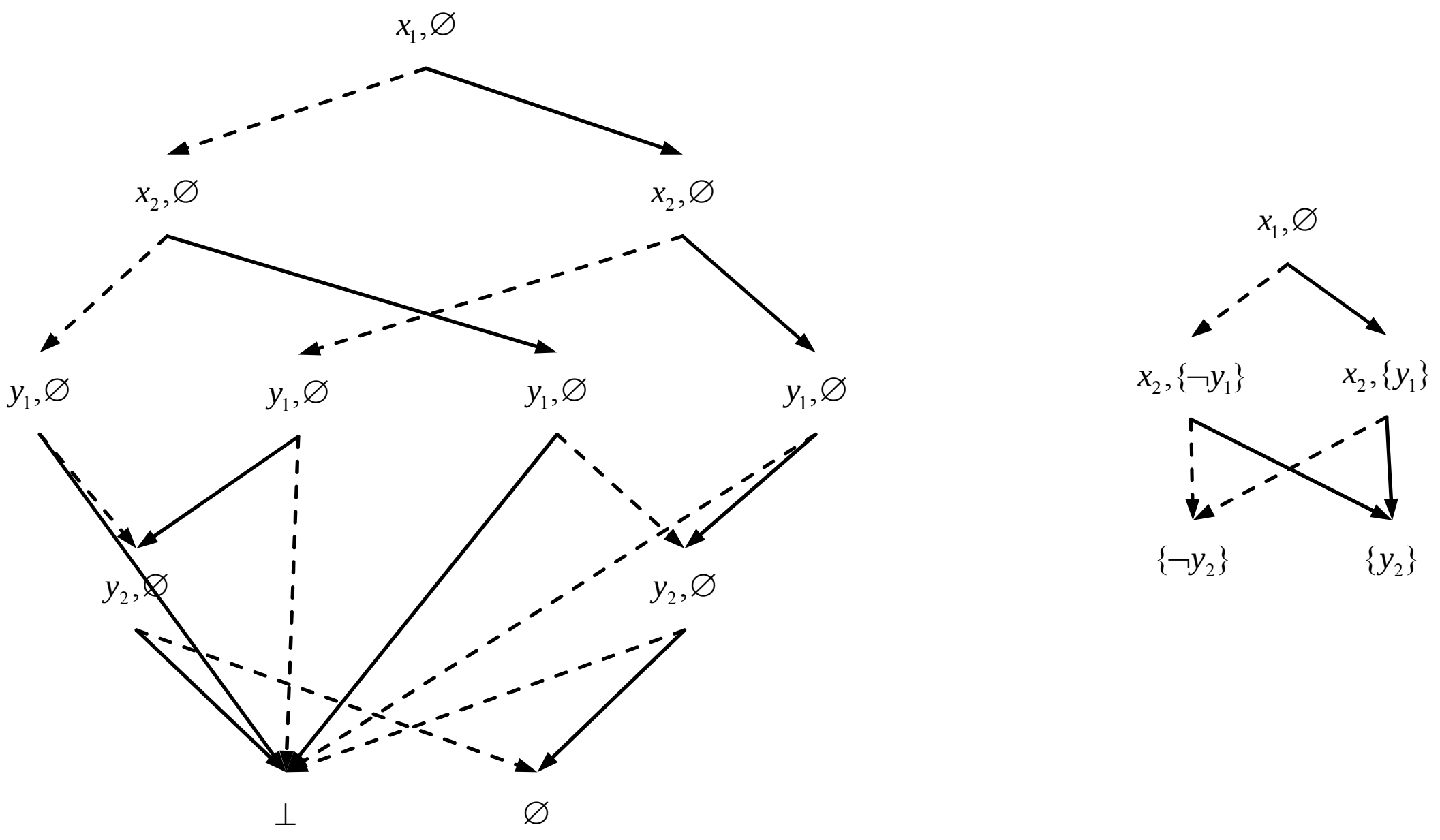

Figure 2. An example about ROBDD-0 (left) and ROBDD-1

It is well known that for any formula there is exactly one ROBDD representing it. We will show that for any $0 \le i \le \infty$, ROBDD-$i$ also has this property. First, we give the definition of Condition [3, 5], which is a useful logical operation in practical applications. Note that we do not restrict this definition to the NNF formulas here.

**Definition 6.** Let $\varphi$ be a formula over the variables set X and let $T$ be a consistent term. The conditioning of $\varphi$ on $T$ (denoted by $\varphi \mid T$, simply denoted by $\varphi \mid l$ if $T = \{l\}$) is a formula obtained by replacing every variable $x$ in $\varphi$ with *true* (resp. *false*) if $x \in T$ (resp. $\neg x \in T$).

Given any formula $\varphi$, variable $x$, literal $l$ and consistent term $T$, by Theorem 1 in [3], we have that: $\varphi \Leftrightarrow (\varphi \mid l) \wedge l$ if $\varphi \Rightarrow l$; $\varphi \Leftrightarrow \varphi \mid x$ if $\varphi \mid x \Leftrightarrow \varphi \mid \neg x$ (we say $x$ can be *omitted* in $\varphi$). Given two equivalent formulas $\varphi_1$ and $\varphi_2$, all variables do not appear in both $\varphi_1$ and $\varphi_2$ can be omitted. And given an OBDD-$L$ with root $v$, according to (*), $\phi(v) \mid L(v) \cup \neg var(x)$ (resp. $\phi(v) \mid L(v) \cup var(x)$) is equivalent to $\phi(lo(v))$ (resp. $\phi(hi(v))$). These observations are used in the proofs of many propositions in this paper, including the following conclusion.

**Proposition 2.** For any formula $\varphi$ over the set of variables $X = \{x_1, \ldots, x_n\}$ with the linear order $x_1 < \ldots < x_n$, and any $0 \le i \le \infty$, there is exactly one ROBDD-$i$ to represent $\varphi$.

**Proof.** By induction on the size of X. The case $|X| = 0$ is obvious. Assume now that we have proven this proposition for $|X| \le n$. We proceed to show it for $|X| = n + 1$.

First we show that there exists some ROBDD-$i$ equivalent to $\varphi$. By the induction hypothesis, this assertion is obvious if there exists some variable $x \in X$ that can be omitted in $\varphi$. Otherwise, we construct a ROBDD-$i$ equivalent to $\varphi$ (In fact, it is just the idea of the compilation algorithm Build, see Section 5). Let $\{l_1, \ldots, l_m\}$ be the set of literals implied by $\varphi$ such that the variable of $l_j$ is less than the one of $l_{j'}$ if $j < j'$, let $L(v) = \{l_1, \ldots, l_{m'}\}$, where $m' = \min\{m, i\}$. If there does not exist any variable appearing in $\varphi$ but not in $L(v)$, then $\langle L(v) \rangle$ is equivalent to $\varphi$. Otherwise let $var(v)$ be the minimum variable appearing in $\varphi$ but not in $L(v)$. It is obvious $\varphi \mid L(v) \cup \{\neg x\}$ is not equivalent to $\varphi \mid L(v) \cup \{x\}$. By the induction hypothesis, the ROBDDs-$i$ corresponding to $\varphi \mid L(v) \cup \{\neg x\}$ and $\varphi \mid L(v) \cup \{x\}$ are not identical to each other, and let $lo(v)$ and $hi(v)$ be their roots, respectively. It is obvious that the DAG with the root $v$ is a ROBDD-$i$ such that $\phi(v)$ is equivalent to $\varphi$.

Then assume that there exist two ROBDDs-$i$ with the roots $u$ and $v$ for $\varphi$, we prove that they are identical to each other. It is obvious that $L(u) = L(v)$ by Proposition 1 and the definition of ROBDD-$i$. And if $var(u) \ne var(v)$, without loss of generality we assume that $var(u) < var(v)$, this means that $var(u)$ can be omitted in $\varphi$, then the ROBDD-$i$ with the root $u$ must have two identical children by the induction hypothesis. This conflicts with the condition 1 of Definition 4. So the formula represented by low (resp. high) child of $v_1$ must be equivalent to the one represented by low (resp. high) child of $v_2$ (otherwise the formula represented by $v_1$ is not equivalent to the one represented by $v_2$), then we know that the low (resp. high) child of $v_1$ is identical to the low (resp. high) child of $v_2$ by induction hypothesis. This means that $v_1$ is identical to $v_2$. ■

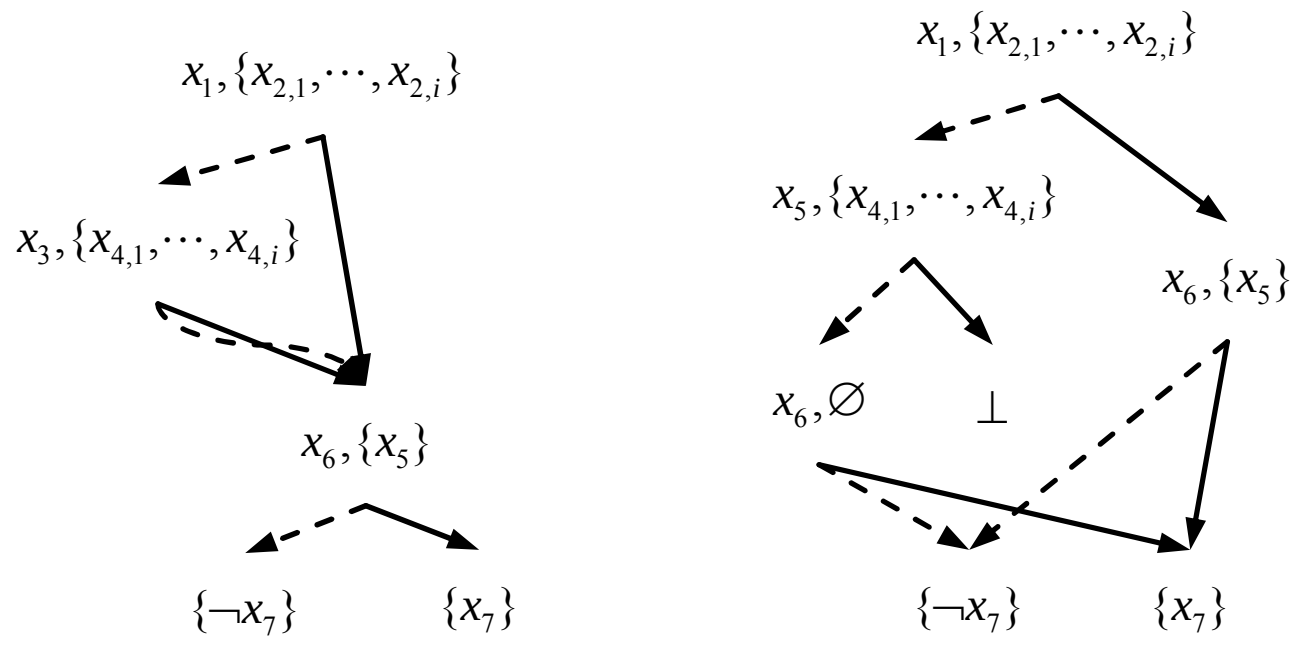

Figure 3. An example about OBDD-$i$ (left) and ROBDD-$i$

Although any ROBDD-$i$ is a reduced OBDD-$i$, there exists some OBDD-$i$ ($0 < i < \infty$) such that the corresponding ROBDD-$i$ has more nodes. An example is showed in Figure 3, where the OBDD-$i$ has 5 nodes and the ROBDD-$i$ has 7 nodes. However, this kind of "strange phenomenon" will not occur when $i = 0$ or $i = \infty$.

**Proposition 3.** Given any formula $\varphi$, the corresponding ROBDD-$i$ has the least number of nodes among all OBDDs-$i$ equivalent to $\varphi$ and each OBDD-$i$ can be transformed into the equivalent ROBDD-$i$ in linear time if $i = 0$ or $i = \infty$.

**Proof.** An algorithm called Reduce which can transform any OBDD-$i$ into the corresponding ROBDD-$i$ in linear time is presented in Figure 4. The function MK is called to guarantee the condition 2 of Definition 4: given any node $u$, if some identical node has appeared before, then MK($v$) returns the old one, otherwise it returns $u$. The condition 1 of Definition 4 is guaranteed by Line 5 (no matter whether $i = \infty$ or not, $L(lo(u)) = \varnothing$ if $lo(u) = hi(u)$, so $L(u)$ does not need to be changed). Every step in a single call (without consideration of the recursive calls) of Reduce can terminal in constant time, include the call of MK, whose running time is constant (readers are referred to [9] for the reason). With the cache $G_1$, there are at most $| v |$ recursive calls of Reduce. So Reduce($v$) can terminate in linear time. And in a single call of Reduce, at most one new node is introduced into the result (i.e. a new node is introduced only when MK($u$) returns $u$ itself). We have that the nodes in the resulting ROBDD-0 or ROBDD-$\infty$ is not more than the ones in the input. Then the conclusion is obvious by Proposition 2. ■

```
procedure Reduce(v)
1:  if G1(v) ≠ empty then return G1(v) endif
2:  Create a new node u
3:  if v is a non-terminal node then
4:      u ← ⟨var(v), Reduce(lo(v)), Reduce(hi(v)), L(v)⟩
5:      if lo(u) = hi(u) then u ← ⟨var(hi(u)), lo(hi(u)), hi(hi(u)), L(v)⟩
6:      endif
7:  endif
8:  G1(v) ← MK(u)
9:  return G1(v)
```

Figure 4. The algorithm Reduce

## 3 On the Succinctness of ROBDD-*L*

In this section, we first show that ROBDD-$\infty$ is the most succinct part in OBDD-$L$, then prove that ROBDD-$i$ is not at least as succinct as ROBDD-$j$ for $i < j$. These mean that ROBDD-$i$ is indeed a different target language from ROBDD-$j$ with $i \neq j$ and ROBDD-$\infty$ is strictly more succinct than any ROBDD-$i$ ($i < \infty$), including ROBDD. Finally, we show that ROBDD-$\infty$ is strictly less succinct than any FBDD. From these results, we can obtain some other conclusions. The definition of succinctness is as follows [5]:

**Definition 7.** Let $L_1$ and $L_2$ be two subsets of NNF. $L_1$ is at least as succinct as $L_2$, if and only if there exists a polynomial $p$ such that for every sentence $\alpha \in L_2$, there exists an equivalent sentence $\beta \in L_1$ where $|\beta| \leq p \cdot |\alpha|$. Here, $|\alpha|$ and $|\beta|$ are the sizes of $\alpha$ and $\beta$, respectively. $L_1$ is strictly more succinct than $L_2$ if and only if $L_1$ is at least as succinct as $L_2$, while $L_2$ is not at least as succinct as $L_1$.

Note that on the one hand the fact $L_1$ is strictly less succinct than $L_2$ does not prevent that $L_1$ is more space efficient than $L_2$ for some kind of knowledge bases, on the other hand the succinctness is only concerned about the best case, in fact for the compilation languages without uniqueness, the time cost of finding the best representation of the knowledge base is so high that usually it is impractical to use such kind of algorithms. Therefore, it is possible that given two languages $L_1$ and $L_2$ such that $L_1$ is strictly more succinct than $L_2$, compared with the $L_2$ compilers, some $L_1$ compiler can generate more space efficient compilation results for some kind of knowledge bases. We will validate this assertion in Section 6.

```
procedure Add-to-inf(v)
1:   function Add-to-inf-sub(v)
2:   |    if G2(v) ≠ empty then return G2(v) endif
3:   |    Create a new node u
4:   |    v′ ← v
5:   |    while v′ is non-terminal and var(v′) does not appear in L∞(v) do
6:   |        if lo(v′) = ⊥ then v′ ← hi(v′)
7:   |        else v′ ← lo(v′)
8:   |        endif
9:   |    endwhile
10:  |    if v′ is a True node then u ← ⟨L∞(v)⟩
11:  |    else
12:  |        var(u) ← var(v′)
13:  |        L(u) ← L∞(v)
14:  |        lo(u) ← Add-to-inf(lo(v′))
15:  |        L(lo(u)) ← L(lo(u)) \ L∞(v)
16:  |        hi(u) ← Add-to-inf(hi(v′))
17:  |        L(hi(u)) ← L(hi(u)) \ L∞(v)
18:  |    endif
19:  |    G2(v) ← u
20:  |    return u
21:  end function Add-to-inf-sub
22:  if v = ⊥ node then return ⊥ endif
23:  For any node v′, compute L∞(v′)
24:  return Reduce(Add-to-inf-sub(v))
```

Figure 5. The algorithm Add-to-inf

**Proposition 4.** ROBDD-∞ is the most succinct subset in OBDD-*L*. In fact, given any OBDD-*L*, let $n$ be the number of the non-terminal nodes such that neither of its children is the False node, then this OBDD-*L* can be transformed into the corresponding ROBDD-∞ in polytime and the nodes of the ROBDD-∞ is not more than $2n + 1$.

**Proof.** We use a specific algorithm to prove this proposition. An algorithm which can transform any OBDD-*L* into the equivalent ROBDD-∞ in polytime is presented in Figure 5. In order to avoid making too many recursive calls, we maintain a cache $G_2$ which stores previously computed outputs of the function Add-to-inf-sub. On Lines 5-9, we search the nearest node $v'$ from $v$ such that neither of the children of $v'$ is the False node. It is obvious that this loop can terminal in polytime.

We first show that the output of Add-to-inf-sub is indeed an OBDD-∞ equivalent to its input such that $VARS(u) \subseteq VARS(v)$, and then Reduce(Add-to-inf-sub($v$)) returns the corresponding ROBDD-∞ by Proposition 3. We prove it by induction on the size of $| v |$. The case $| v | = 1$ is immediate. Assume that the conclusion holds for $| v | \le n$. When $| v | = n + 1$, the output of Add-to-inf-sub obviously meets the requirements if $v'$ is a True node. Otherwise we have $| lo(v') | \le n$ and $| hi(v') | \le n$. By the induction hypothesis, the output of Add-to-inf-sub($lo(v')$) (resp. Add-to-inf-sub($hi(v')$)) is an OBDD-∞ equivalent to $\phi(lo(v'))$ (resp. $\phi(hi(v'))$). After Line 15 (resp. Line 17), the BDD-*L* with the root $lo(u)$ (resp. $hi(u)$) is still an OBDD-∞. Given any literal $l \in L(u)$ and $x$ is the variable of $l$, if $l$ appears in $v'$ or some ancestor node of $v'$, and then $x$ doesn't appear in any descendent node of $v'$, we have that $x$ doesn't appear in any descendent node of $u$ because $VARS(lo(u)) \subseteq VARS(lo(v'))$ and $VARS(hi(u)) \subseteq VARS(hi(v'))$ by the induction hypothesis. Otherwise $l \in L_\infty(lo(v'))$ and $l \in L_\infty(hi(v'))$ by Definition 2, then by Proposition 1 and the induction hypothesis, $l \in L$(Add-to-inf-sub($lo(v')$)) and $l \in L$(Add-to-inf-sub($hi(v')$)), then $l$ is deleted on Lines 15 and 17, it means that $x$ doesn't appear in any descendent node $u$. So the output of Add-to-inf-sub($v$) is a BDD-*L*. By Definition 3, $var(v')$ is less than any variable appearing in $lo(v')$ and $hi(v')$, and then $var(v')$ is less than any variable appearing in $lo(u)$ and $hi(u)$ because $VARS(lo(u)) \subseteq VARS(lo(v'))$ and $VARS(hi(u)) \subseteq VARS(hi(v'))$. We know that $L(lo(u))$ does not share any literal with $L(hi(u))$ and neither of $lo(u)$ and $hi(u)$ is the False node, this means that $L(u) = L_\infty(u)$ by Definition 2, and then the condition 3 of Definition 3 is satisfied in $v$. So the output of Add-to-inf-sub($v$) is an OBDD-∞.

Then we show that the call of Add-to-inf-sub($v$) can terminate in polytime for any ROBDD-*L* with the root $v$, then Add-to-inf($v$) can terminate in polytime because we can compute $L_\infty(v')$ for any node $v'$ in polytime (Line 23) and the output can be transformed into the equivalent ROBDD-∞ in polytime by Proposition 3 on Line 24. It is obvious that a single call of Add-to-inf-sub can terminate in polytime because every step except the recursive calls can terminal in polytime. With the cache $G_2$, there are at most $n$ recursive calls of Add-to-inf-sub. So Add-to-inf-sub($v$) can terminate in polytime.

Finally, in a single call of Add-to-inf-sub, exactly two new nodes (i.e., $lo(u)$ and $hi(u)$) will be introduced into the resulting OBDD-∞. Together with the root, there exist $2n + 1$ nodes in the resulting OBDD-∞. By Proposition 3, there exist at most $2n + 1$ nodes in the resulting ROBDD-∞. In summary, the conclusion holds. ■

It is pointed out that the algorithm Add-to-inf immediately give us a ROBDD-∞ compilation method, i.e., firstly compile the knowledge base into ROBDD, and then use the algorithm Add-to-inf to turn the result into ROBDD-∞. In addition, we point out one observation about Add-to-inf as follows:

Observation 1. If the input of Add-to-inf-sub satisfies the formula represented by the low child of any node is not equivalent to the high child, then its output also satisfies this condition. We can prove it by induction. So we only need to delete the redundant nodes from the output of Add-to-inf-sub in the algorithm Reduce.

As the algorithm Add-to-inf plays a key role in this paper, we give an example to show how it works.

**Example 1.** Let us consider the ROBDD-0 in Figure 2. For simplicity, we omit the implied literals table here, and the nodes table of ROBDD is showed in Figure 6(a). First, we compute the maximal set of implied literals for each node in ROBDD. Then Add-to-inf-sub($v_{10}$) is called. As both children of $v_{10}$ are non-False, Add-to-inf-sub($v_9$) is called on the Line 14. Similarly, Add-to-inf-sub($v_7$) is called on the Line 14. After running Lines 5-9, $v' = v_1$. Then $u$ is assigned as $\langle\{\neg y_1, \neg y_2\}\rangle$ and it is put into $G_2$. $L(lo(u))$ is assigned as $\{\neg y_2\}$ on Line 15, this means that $v_{\text{IV}}$ is generated. Then Add-to-inf-sub($v_5$) is called on the Line 16. Similarly, $\langle\{\neg y_1, y_2\}\rangle$ and it is put into $G_2$, $L(hi(u))$ is assigned as $\{y_2\}$ on Line 17, this means that $v_{\text{II}}$ is generated. Then $\langle x_2, \text{IV}, \text{II}, \{\neg y_1\}\rangle$ is put into $G_2$. Back to the calling of Add-to-inf-sub($v_{10}$), $L(lo(u))$ is assigned as $\{\neg y_1\}$ on Line 15, this means that $v_{\text{VI}}$ is generated. Similarly, $L(hi(u))$ is assigned as $\{y_1\}$ on Line 17 and $v_{\text{V}}$ is generated. Finally, Add-to-inf-sub($v_{10}$) returns $\langle x_1, \text{VI}, \text{V}, \varnothing\rangle$, which is the root of the resulting OBDD-∞ (its nodes table is showed in Figure 6(c), and it is showed in Figure 6(d)). On Lines 24, we run Reduce on the OBDD-∞, and the ROBDD-∞ in Figure 2 is generated.

(a) The nodes table of ROBDD-0

| ID | *var* | *low* | *high* | *L* | $L_\infty$ |
|---|---|---|---|---|---|
| 10 | $x_1$ | 9 | 8 | ∅ | ∅ |
| 9 | $x_2$ | 7 | 5 | ∅ | $\{\neg y_1\}$ |
| 8 | $x_2$ | 6 | 4 | ∅ | $\{y_1\}$ |
| 7 | $y_1$ | 3 | 0 | ∅ | $\{\neg y_1, \neg y_2\}$ |
| 6 | $y_1$ | 0 | 3 | ∅ | $\{y_1, \neg y_2\}$ |
| 5 | $y_1$ | 2 | 0 | ∅ | $\{\neg y_1, y_2\}$ |
| 4 | $y_1$ | 0 | 2 | ∅ | $\{y_1, y_2\}$ |
| 3 | $y_2$ | 1 | 0 | ∅ | $\{\neg y_2\}$ |
| 2 | $y_2$ | 0 | 1 | ∅ | $\{y_2\}$ |
| 1 | True | - | - | ∅ | ∅ |
| 0 | False | - | - | - | - |

(b) The cache $G_2$

| ID | $G_2$ | | | |
|---|---|---|---|---|
| | *var* | *low* | *high* | *L* |
| 10 | $x_1$ | VI | V | ∅ |
| 9 | $x_2$ | IV | II | $\{\neg y_1\}$ |
| 8 | $x_2$ | III | I | $\{y_1\}$ |
| 7 | True | - | - | $\{\neg y_1, \neg y_2\}$ |
| 6 | True | - | - | $\{y_1, \neg y_2\}$ |
| 5 | True | - | - | $\{\neg y_1, y_2\}$ |
| 4 | True | - | - | $\{y_1, y_2\}$ |

(c) The nodes table of OBDD-∞

| ID | *var* | *low* | *high* | *L* |
|---|---|---|---|---|
| VII | $y_1$ | VI | V | ∅ |
| VI | $y_2$ | IV | II | $\{\neg y_1\}$ |
| V | $y_2$ | III | I | $\{y_1\}$ |
| IV | $y_2$ | - | - | $\{\neg y_2\}$ |
| III | $y_2$ | - | - | $\{\neg y_2\}$ |
| II | $y_2$ | - | - | $\{y_2\}$ |
| I | $y_2$ | - | - | $\{y_2\}$ |

(d) The resulting OBDD-∞

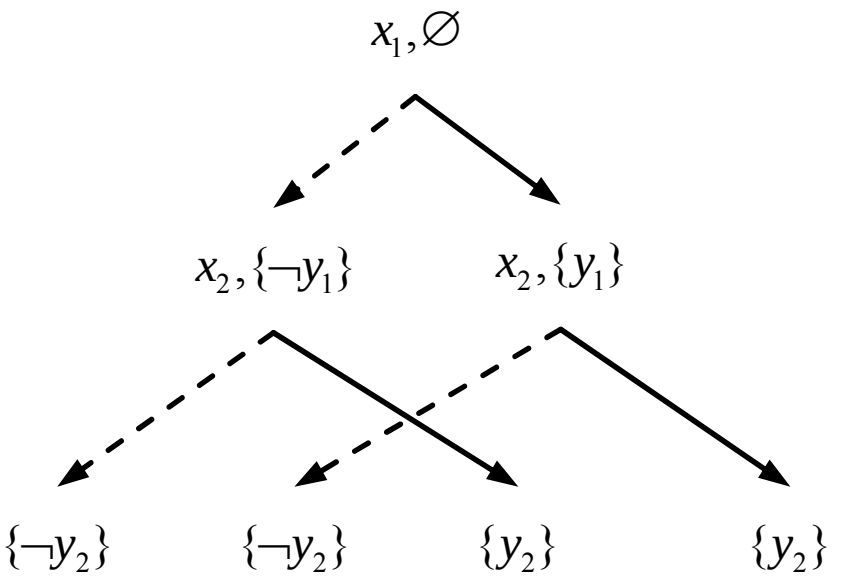

Figure 6. An example about the algorithm Add-to-inf.

**Proposition 5.** Given any two number $i < j$, ROBDD-$i$ is not at least as succinct as ROBDD-$j$.

**Proof.** According to Definition 7, we just need to provide a counterexample here. Let us consider the formula below over the linear order $x_1 < \ldots < x_{i+1} < x_{1,1} < \ldots < x_{1,i+1} < \ldots < x_{n,1} < \ldots < x_{n,i+1}$:

$$((x_1 \leftrightarrow x_{1,1}) \wedge \cdots \wedge (x_1 \leftrightarrow x_{1,i+1})) \wedge \cdots \wedge ((x_n \leftrightarrow x_{n,1}) \wedge \cdots \wedge (x_n \leftrightarrow x_{n,i+1})).$$

The size of ROBDD-$i$ representing this formula is exponential, while the size of ROBDD-$(i + 1)$ is polynomial. And ROBDD-$(i + 1)$ is as same as ROBDD-$j$. ■

Now we know that ROBDD-∞ is strictly more succinct than ROBDD by the above two propositions, then a question occurs: is it possible that ROBDD-∞ has a linear size as ROBDD for some kind of knowledge bases. The following proposition, which is useful for proving that the size of ROBDD-$i$ ($0 < i \leq \infty$) corresponding to some formula $\varphi$ is exponential in the size of $\varphi$, will answer this question.

**Proposition 6.** Given any formula $\varphi$ over the variables $x_1 < \ldots < x_n$ and $0 < i \le \infty$, if $\varphi \Rightarrow C$ does not hold for any non-tautology clause such that $x_n$ does not appear in it, then we have that: all of the ROBDD-*i* corresponding to $\varphi$ are the same; assume that ROBDD-*i* has the root $v$ and ROBDD has the root $u$, then

$$\left\lceil \frac{|u|-1}{2} \right\rceil \le |v| \le 2 \cdot |u| - 5 .$$

**Proof.** The case $\varphi = false$ is impossible. Given any non-terminal node $v_1$ in ROBDD-1, we have that $L_\infty(v_1)$ has no other literal except $x_n$ and $\neg x_n$, otherwise there exists some literal $l$ such that $l \notin \{x_n, \neg x_n\}$ and $\varphi \wedge path(v_1) \Rightarrow l$ by Proposition 1, and then $\varphi \Rightarrow \neg path(v_1) \vee l$, it is obvious that $\neg path(v_1) \vee l$ is a non-tautology clause and $x_n$ does not appear in it, this is impossible. So there does not exist some node $v_2$ such that $var(v_2) = x_n$, otherwise we have that: any father $v_3$ of $v_2$ satisfies $L_\infty(v_3) = \varnothing$ by Definition 3 and Proposition 2; the children of $v_2$ are $\bot$ and $\langle\varnothing\rangle$ by Definition 4. These two assertions conflict with each other. Then given any True node $v_4$, $L_\infty(v_4)$ has no other literal except $x_n$ and $\neg x_n$ for the same reason as $v_1$. In summary, given any node $v'$ in ROBDD-1, $|L_\infty(v_1)| \le 1$. Then by Definition 5, all of the ROBDD-*i* corresponding to $\varphi$ are the same.

We call Add-to-inf($u$) to transform the ROBDD into the ROBDD-$\infty$. Given any non-terminal node $u_1$ in ROBDD with $var(u_1) < x_n$, neither of its children is the False node, otherwise, without loss of generality, assume that $lo(u_1) = \bot$, $\varphi \wedge path(u_1) \Rightarrow var(u_1)$, and then $\varphi \Rightarrow \neg path(u_1) \vee var(u_1)$, it is obvious that $\neg path(u_1) \vee var(u_1)$ is a non-tautology clause and $x_n$ does not appear in it, this is impossible. So there exists some path from $u_1$ to $u_2$ such that $var(u_2) = x_n$, otherwise $\phi(u_1) \Leftrightarrow true$. Obviously, one child of $u_2$ is $\bot$, while other child is $\langle\varnothing\rangle$. So $x_n$ cannot be omitted in any non-terminal node in ROBDD. Then the number of non-terminal nodes such that neither of its children is the False node in ROBDD is not more than $|u| - 3$. We have that $|v| \le 2 \cdot (|u| - 3) + 1 = 2 \cdot |u| - 5$ by Proposition 4.

Let non-False node $u_3$ be one child of $u_1$, the new node created in the single call of Add-to-inf-sub($u_1$) must be equivalent to $\phi(u_3)$ if $L_\infty(u_3)$ is empty, the new node must be equivalent to $\phi(u_3)$ or $\phi(u_3) \mid x_n$ if $x_n \in L_\infty(u_3)$, the new node must be equivalent to $\phi(u_3)$ or $\phi(u_3) \mid \neg x_n$ otherwise. Let $\Phi_1 = \{\phi(u_3)\colon L_\infty(u_3) = \varnothing, u_3 \ne \langle\varnothing\rangle\}$, $\Phi_2 = \{\phi(u_3)\colon x_n \in L_\infty(u_3)\}$, $\Phi_3 = \{\phi(u_3) \mid x_n\colon x_n \in L_\infty(u_3)\}$, $\Phi_4 = \{\phi(u_3)\colon \neg x_n \in L_\infty(u_3)\}$, $\Phi_5 = \{\phi(u_3) \mid \neg x_n\colon \neg x_n \in L_\infty(u_3)\}$. Obviously, the elements in $\Phi_j$ $(1 \le j \le 5)$ are not equivalent to each other, the elements in $\Phi_1 \cup \Phi_2 \cup \Phi_4$ are not equivalent to each other, $|\Phi_1 \cup \Phi_2 \cup \Phi_4| = |\Phi_1| + |\Phi_2| + |\Phi_4| = |u| - 3$ (without the root $u$ and the terminal nodes). Given any formula $\varphi_1 \in \Phi_1 \cup \Phi_2 \cup \Phi_3 \cup \Phi_4 \cup \Phi_5$ without the appearance of $x_n$, $\varphi_1$ could only appears in $\Phi_1$, $\Phi_3$ and $\Phi_5$ if $\varphi_1 = true$, otherwise $\varphi_1$ could only be equivalent to some formula in $\Phi_3$ and $\Phi_5$ because $x_n$ cannot be omitted in any non-terminal node in ROBDD. Let $\Phi_6 = \{\varphi_1\colon \varphi_1 \in \Phi_3$ and $\varphi_1$ is equivalent to some formula in $\Phi_5\}$. It is obvious that

$$|\Phi_6| \le \min(|\Phi_3|, |\Phi_5|) = \min(|\Phi_2|, |\Phi_4|) \le \left\lceil \frac{|u|-3}{2} \right\rceil$$

and each element in ROBDD corresponds to at least one node in the output of Add-to-inf-sub($u$). By Observation 1, we only need to delete redundant nodes from the output of Add-to-inf-sub($u$) because the input is an ROBDD, then we have that

$$|v| \ge 1 + (|u| - 3) - |\Phi_6| \ge 1 + \left\lceil \frac{|u|-3}{2} \right\rceil = \left\lceil \frac{|u|-1}{2} \right\rceil . \qquad \blacksquare$$

Now we turn to show that ROBDD-$\infty$ is strictly less succinct than FBDD:

**Proposition 7.** ROBDD-$\infty$ is strictly less succinct than FBDD.

**Proof**. First, we use a specific algorithm to show that FBDD is as succinct as ROBDD-$\infty$. An algorithm which can transform any ROBDD-$\infty$ into the equivalent FBDD in polytime is presented in Figure 7. Here we treat FBDD as a subset of BDD-0, i.e., for any non-False in FBDD, $L(v) = \varnothing$. By induction, it is easy to prove that Inf2FBDD is correct.

```
Inf2FBDD(v)
    if G3(v) ≠ empty then return G3(v) endif
    Create a FBDD node u
    if v = ⊥ then u ← ⊥; MK(u)
    else
        if v is a non-terminal node then
            u ← ⟨var(v), Inf2FBDD(lo(v)), Inf2FBDD(hi(v)), ∅⟩
        else u ← ⟨∅⟩
        endif
        MK(u)
        for any x ∈ L(v) do u ← ⟨x, ⊥, u, ∅⟩; MK(u) endfor
        for any ¬x ∈ L(v) do u ← ⟨x, u, ⊥, ∅⟩; MK(u) endfor
    endif
    G3(v) ← u
    return u
```

Figure 7. The algorithm Inf2FBDD

Then we show that ROBDD-$\infty$ is not as succinct as FBDD by given a counterexample. Let us consider the formula $\varphi = (x_1 \leftrightarrow y_1) \vee \ldots \vee (x_n \leftrightarrow y_n)$ over the variables order $x_1 < \ldots < x_n < y_1 < \ldots < y_n$. It is well known that there exists a linear FBDD equivalent to $\varphi$ and the size of ROBDD representing $\varphi$ is exponential. Obviously, $\varphi$ does not imply any non-tautology clause without the appearance of $y_n$. By Proposition 6, the size of ROBDD-∞ representing $\varphi$ is exponential. So the conclusion holds. ■

By Propositions 4, 5, 7, Proposition 3.1 in [5], and the transitivity of succinctness, we know that:

**Corollary 1.** ROBDD-∞ is strictly less succinct than DNNF and d-DNNF; ROBDD-∞ is incomparable to DNF, CNF, IP and PI.

## 4 The Operations of ROBDD-∞

We have showed that ROBDD-∞ is the most succinct subset in OBDD-*L* in the previous section. In order to evaluate the inferential power of this interesting subset, we analyze the operations that ROBDD-∞ supports in polytime with respect to the criterion proposed in the knowledge map in this section. The following queries and transformations have been considered in the knowledge map. We just recall them here and readers are referred to [5] for their importance.

**Definition 8**. Given any subset L of NNF,

L satisfies **CO** (resp. **VA**) iff there exists a polytime algorithm that maps every formula $\varphi$ from L to 1 if $\varphi$ is consistent (resp. valid), and to 0 otherwise;

L satisfies **CE** iff there exists a polytime algorithm that maps every formula $\varphi$ from L and every clause $C$ to 1 if $\varphi \Rightarrow C$ holds, and to 0 otherwise;

L satisfies **IM** iff there exists a polytime algorithm that maps every formula $\varphi$ from L and every term $T$ to 1 if $T \Rightarrow \varphi$ holds, and to 0 otherwise;

L satisfies **EQ** (resp. **SE**) iff there exists a polytime algorithm that maps every pair of formulas $\varphi$, $\gamma$ from L to 1 if $\varphi \Leftrightarrow \gamma$ (resp. $\varphi \Rightarrow \gamma$) holds, and to 0 otherwise;

L satisfies **CT** iff there exists a polytime algorithm that maps every formula $\varphi$ from L and some the variables set X which includes all of the variables appearing in $\varphi$ to a non-negative integer that represents the number of models of $\varphi$ over X (in binary notation).

L satisfies **ME** iff there exists a polynomial $p(.,.)$ and an algorithm that outputs all models of an arbitrary formula $\varphi$ from L over some the variables set X which includes all of the variables appearing in $\varphi$ in time $p(n, m)$, where $n$ is the size of $\varphi$ and m is the number of its models over X.

**Definition 9**. Given any subset L of NNF,

L satisfies **CD** iff there exists a polytime algorithm that maps every formula $\varphi$ from L and every consistent term $T$ to a formula from L that is logically equivalent to $\varphi \mid T$.

L satisfies **FO** iff there exists a polytime algorithm that maps every formula $\varphi$ from L and every subset X of the set of variables appearing in $\varphi$ to a formula from L that is equivalent to $\exists X.\varphi$, i.e. the formula that does not mention any variable from X and for every formula $\gamma$ that does not mention any variable from X, we have $\varphi \Rightarrow \gamma$ precisely when $\exists X.\varphi \Rightarrow \gamma$. If the property holds for singleton X, we say that L satisfies **SFO**.

L satisfies $\wedge$**C** (resp. $\vee$**C**) iff there exists a polytime algorithm that maps every finite set of formulas $\varphi_1, \ldots, \varphi_n$ from L to a formula of L that is logically equivalent to $\varphi_1 \wedge \ldots \wedge \varphi_n$ (resp. $\varphi_1 \vee \ldots \vee \varphi_n$).

L satisfies $\wedge$**BC** (resp. $\vee$**BC**) iff there exists a polytime algorithm that maps every pair of formulas $\varphi$ and $\gamma$ from L to a formula of L that is logically equivalent to $\varphi \wedge \gamma$ (resp. $\varphi \vee \gamma$).

L satisfies $\neg$**C** iff there exists a polytime algorithm that maps every formula $\varphi$ from L to a formula of L that is logically equivalent to $\neg\varphi$.

Table 1. The polytime query of ROBDD-∞. √ means "satisfies", ∘ means "does not satisfy unless P = NP", and ? means "unknown".

| L | **CO** | **VA** | **CE** | **IM** | **EQ** | **SE** | **CT** | **ME** |
|---|---|---|---|---|---|---|---|---|
| ROBDD | √ | √ | √ | √ | √ | √ | √ | √ |
| ROBDD-∞ | √ | √ | √ | √ | √ | ? | √ | √ |
| FBDD | √ | √ | √ | √ | ? | ∘ | √ | √ |
| d-DNNF | √ | √ | √ | √ | ? | ∘ | √ | √ |

Table 1 summarizes query-related properties of ROBDD-∞. As ROBDD, FBDD and d-DNNF are three of the most widely used target languages in practical applications, their properties are also showed here for comparison.

**Proposition 8**. The results in Table 1 hold.

**Proof**. **CO**, **VA** and **EQ**: Recall that for any formula, there is exactly one ROBDD-∞ representing it. This means, in particular, that there is exactly one ROBDD-∞ for the constant formula *true* (resp. *false*): the True node with no implied literal (resp. False node). And given two ROBDDs-∞, they are equivalent if and only if they are the same. So we have that: (1) Deciding the satisfiability and validity of a ROBDD-∞ can be done in constant time, it means that **CO** and **VA** are satisfied; (2) Deciding the equivalence between two ROBDDs-∞ can be done in polytime, it means that **EQ** is satisfied.

**CT**, **CE**, **IM** and **ME**: Counting the models of a ROBDD-∞ over a set of variables can be done in linear time. An algorithm is presented in Figure 8. A single call of Count can terminal in constant time with the data unit size in implied literals table. With the cache $G_4$, the number of recursive calls is $|v|$. Then this algorithm has a linear time complexity. Furthermore, we prove that the result of Count($v$, X) equals to the number of models of $\phi(v)$ over X by induction on the size of $|v|$. The case $|v| = 1$ is immediate. Assume that Count($v$, X) returns the correct result for $|v| \le n$. When $|v| = n + 1$, it is obvious that $|lo(v)| \le n$ and $|hi(v)| \le n$. According to (*), the number of models of $\phi(v)$ equals to the sum of number of models of $(\bigwedge l \in L(v)) \wedge \neg var(v) \wedge \phi(lo(v))$ and $(\bigwedge l \in L(v)) \wedge var(v) \wedge \phi(hi(v))$. By the induction hypothesis and Definition 3, and the number of models of $(\bigwedge l \in L(v)) \wedge \neg var(v) \wedge \phi(lo(v))$ (resp. $(\bigwedge l \in L(v)) \wedge var(v) \wedge \phi(hi(v))$) equals to Count($lo(v)$, X) / $2^{|L(v)|+1}$ (resp. Count($hi(v)$, X) / $2^{|L(v)|+1}$). So the induction hypothesis holds. It means that **CT** is satisfied. In Proposition 9, we will show that ROBDD-∞ satisfies **CD**. Then by Lemma A.3, A.4 and A.7 in [5], we know that ROBDD-∞ satisfies **ME**, **CE** and **IM**. ■

```
procedure Count(v, X)
    if G4(v) ≠ empty then return G4(v) endif
    if v is a terminal node then
        if v = ⊥ then G4(v) ← 0
        else G4(v) ← 2^(|X| − |L(v)|)
        endif
    else G4(v) ← (Count(lo(v), X) + Count(hi(v), X)) / 2^(|L(v)| + 1)
    endif
    return G4(v)
```

Figure 8. The algorithm Count

We know that two broad areas in formal verification of hardware are distinguished: checking whether a combinational circuit complies with a given specification; checking whether a circuit's behavior conforms to certain desired properties. The former is a case of equivalence checking, while the latter is mostly a case of clausal entailment. So ROBDD-∞ is potential to be widely used in verification field.

Note that we just need some fragment information (i.e., the size) of the implied literals in Count and we pre-record them in implied literals table so that we can visit a node of ROBDD-∞ in constant time. The same situation occurs in computing the minimum cardinality [3], which is useful in model-based diagnose, we only need an extra data unit in implied literals table to pre-record the number of negative literals in implied literals set. By Proposition 4, every OBDD-*L* can be transformed into the corresponding ROBDD-∞ in polytime, and then we have that:

**Corollary 2.** OBDD-*L* satisfies **CO**, **VA**, **CE**, **IM**, **EQ**, **CT** and **ME**.

Table 2. The polytime transformations of ROBDD-∞. √ means "satisfies", • means "does not satisfy", ∘ means "does not satisfy unless P = NP", and ? means "unknown".

| L | **CD** | **FO** | **SFO** | ∧**C** | ∧**BC** | ∨**C** | ∨**BC** | ¬**C** |
|---|---|---|---|---|---|---|---|---|
| ROBDD | √ | • | √ | • | √ | • | √ | √ |
| ROBDD-∞ | √ | • | • | • | ? | • | • | • |
| FBDD | √ | • | ∘ | • | ∘ | • | ∘ | √ |
| d-DNNF | √ | ∘ | ∘ | ∘ | ∘ | ∘ | ∘ | ? |

Table 2 summarizes transformation-related properties of ROBDD-∞. Again, Properties of d-DNNF, FBDD and ROBDD are also showed here for comparison.

**Proposition 9**. The results in Table 2 hold.

**Proof**. **CD**: An algorithm is presented in Figure 9. With the cache $G_5$, this function Condition-sub will terminate in polytime. And we have showed that Add-to-inf can terminate in polytime in the previous section. So the algorithm Condition has a polynomial time complexity. Moreover, we prove that the result of Condition-sub($v$, $T$) is an OBDD-*L* equivalent to $\phi(v) \mid T$ by induction on the size of $|v|$. The case $|v| = 1$ is immediate. Assume that the output of Condition-sub($v$, $T$) is an OBDD-*L* equivalent to $\phi(v) \mid T$ and $VARS(u) \subseteq VARS(v)$ for $|v| \le n$. When $|v| = n + 1$, it is obvious that $|lo(v)| \le n$ and $|hi(v)| \le n$. By (*) and Definition 6, we have that

$$\phi(v) \mid T = (\bigwedge l \in L(v)) \mid T \wedge ((\neg var(v) \mid T \wedge \phi(lo(v)) \mid T) \vee (var(v) \mid T \wedge \phi(hi(v)) \mid T)).$$

Line 13 guarantees that at least one child of $u$ is non-False if $u$ is non-terminal. Then by the induction hypothesis and Definition 3, the DAG with the root $u$ is an OBDD-*L* such that $\phi(u) \Leftrightarrow \phi(v) \mid T$ and $VARS(u) \subseteq VARS(v)$. Note that $L(v) \setminus T$ and $L(v) \cap \neg T \ne \emptyset$ are needed to be computed just once for the same sets of implied literals set with the use of implied literals table. This means that the implied literals table is helpful for saving the computing time of the logical operations such as conditioning, clausal entailment and so on.

```
procedure Condition(v, T)
    function Condition-sub(v, T)
        if G5(v) ≠ empty then return G5(v) endif
        Create a new node u
        if L(v) ∩ ¬T ≠ ∅ then u ← ⊥
        else-if v is a True node then u ← ⟨L(v) \ T⟩
        else
            if var(v) ⊆ T then u1 ← ⊥
            else u2 ← Condition-sub(lo(v), T)
            endif
            if ¬var(v) ⊆ T then u2 ← ⊥
            else u2 ← Condition-sub(hi(v), T)
            endif
            if both u1 and u2 are ⊥ then u ← ⊥
            else u ← (var(v), u1, u2, L(v) \ T)
            endif
        endif
        MK(u)
        G5(v) ← u
        return u
    end function Condition-sub
    if v = ⊥ then return ⊥ endif
    return Add-to-inf(Condition-sub(v, T))
```

Figure 9. The algorithm Condition

$\vee$**BC**, $\vee$**C**, **SFO** and **FO**: Check the following formula over the variables order $x_1 < \ldots < x_n < y_1 < \ldots < y_n < z$:

$$\varphi \vee \gamma = ((x_1 \leftrightarrow y_1) \wedge \cdots \wedge (x_n \leftrightarrow y_n)) \vee z$$

The ROBDD-∞ corresponding to $\varphi$ with the root $v$ has a linear size and the ROBDD-∞ corresponding to $\gamma$ only has the node $\langle\{z\}\rangle$. The ROBDD-0 corresponding to $\varphi \vee \gamma$ is obviously exponential and $\varphi \vee \gamma$ satisfies the condition in Proposition 6, so the size of the corresponding ROBDD-∞ is also exponential. Therefore, there exists no polytime algorithm which can map the ROBDDs-∞ corresponding to $\varphi$ and $\gamma$ to the ROBDD-∞ corresponding to $\varphi \vee \gamma$. This means that $\vee$**BC** cannot be satisfied. Then $\vee$**C** also cannot be satisfied. Let $var(u) = x$, $L(u) = \varnothing$, $lo(u) = v$ and $hi(u) = \langle\{z\}\rangle$. Obviously, $\phi(u) \Leftrightarrow (x \wedge \varphi) \vee (\neg x \wedge \gamma)$. So there exists no polytime algorithm which can map the ROBDD-∞ with the root $u$ to the ROBDD-∞ corresponding to $\exists x.\phi(u)$ because $\varphi \vee \gamma \Leftrightarrow \exists x.((x \wedge \varphi) \vee (\neg x \wedge \gamma))$. This means that **SFO** cannot be satisfied. Then **FO** also cannot be satisfied.

$\wedge$**C**: By Corollary 1, ROBDD-∞ is not at least as succinct as CNF, and then there exists some CNF formula $\varphi$ that cannot be transformed into the corresponding ROBDD-∞ in polytime. Obviously, each clause in $\varphi$ can be transformed into the corresponding ROBDD-∞ in linear time. If ROBDD-∞ satisfies $\wedge$**C**, then $\varphi$ can be transformed into the corresponding ROBDD-∞ in polytime. This is impossible. So $\wedge$**C** cannot be satisfied.

$\neg$**C**: Here is a counterexample over the variables order $x_1 < \ldots < x_n < y_1 < \ldots < y_n$:

$$\varphi = ((x_1 \leftrightarrow y_1) \wedge \cdots \wedge (x_n \leftrightarrow y_n))$$

Obviously, $\neg\varphi$ satisfies the condition in Proposition 6 and the size of the ROBDD corresponding to it is exponential, so the size of the corresponding ROBDD-∞ is also exponential. However, the size of the ROBDD-∞ corresponding to $\varphi$ is linear. So there exists no polytime algorithm which can map the ROBDD-∞ corresponding to $\varphi$ to the ROBDD-∞ corresponding to $\neg\varphi$. This means that $\neg$**C** cannot be satisfied. ■

We close this section by a detailed theoretical comparison between ROBDD, ROBDD-∞, FBDD and d-DNNF. ROBDD and ROBDD-∞ have the uniqueness over a specific variables order, while neither FBDD nor d-DNNF has this property. ROBDD-∞ is strictly more succinct than ROBDD, but it is strictly less succinct than FBDD, which is strictly less succinct than d-DNNF. From Table 1, ROBDD satisfies **SE** which ROBDD-∞ does not satisfy, while ROBDD-∞ satisfies **EQ** which neither FBDD nor d-DNNF satisfies. In the future, practical applications might need some other class of queries (i.e., not involved in the knowledge compilation map) such that ROBDD can answer them in polytime, while none of ROBDD-∞, FBDD and d-DNNF can, or both ROBDD and ROBDD-∞ can answer them in polytime, while neither FBDD nor d-DNNF can, or only d-DNNF cannot. From Table 2, it seems that both d-DNNF and ROBDD-∞ have some problems for the transformations, while FBDD is a bit better than them. Fortunately, the transformations usually are completed in the off-line phase.

## 5 Compiling Propositional Theory into ROBDD-*i*

In this section we present two compilation algorithms about ROBDD-$i$ ($0 \leq i \leq \infty$). We focus on top-down compilation algorithms rather than bottom-up ones for the reasons that: we do not know whether ROBDD-∞ satisfies $\wedge$**BC** or not; a well-known problem with the bottom-up methods is that the intermediate results that arise in the process can grow so large as to make further manipulation impossible, even when the final result would have a tractable size [20] (in the extreme case when the knowledge base is unsatisfiable, the final compilation result of ROBDD-$i$ will has only one node, but the intermediate results may be satisfiable and have many nodes). First, we propose a compilation algorithm called Build for ROBDD-$i$ ($0 \leq i \leq \infty$). Then we optimize it by the use of some specific models of propositional theory and devise another algorithm called Build-inf for ROBDD-$\infty$. Finally, we introduce some techniques which are potential to improve the performance of Build and Build-inf when the input is in CNF.

### 5.1 Compilation Algorithm

The compilation algorithm Build is presented in Figure 10. Its correctness is guaranteed by the following proposition.

**Proposition 10**. Given any propositional formula $\varphi$, Build($\varphi$, $i$) can terminate in finite time and its output is the corresponding ROBDD-$i$.

**Proof**. Let X be the set of variables appearing in $\varphi$. We prove the proposition by induction on the size of $|X|$. Assume that Build($\varphi$, $i$) can terminate in finite time and its output is the ROBDD-$i$ corresponding to $\varphi$ with the root $v$ such that $VARS(v) \subseteq X$ if $|X| \le n$. The case $|X| = 0$ is immediate. When $|X| = n + 1$, we proceed by case analysis:

$\varphi$ is unsatisfiable: The conclusion is obvious.

$\varphi$ is equivalent to *true* after Line 9: The conclusion is obvious.

Otherwise: After Line 9, let $X_1$ and $X_2$ be the set of variables appearing in $\varphi \mid \neg x_j$ and $\varphi \mid x_j$, respectively. By the induction hypothesis, Build($\varphi \mid \neg x_j$, $i$) (resp. Build($\varphi \mid x_j$, $i$)) can terminate in finite time and its output is the ROBDD-$i$ corresponding to $\varphi \mid \neg x_j$ (resp. $\varphi \mid x_j$) such that $VARS(lo(v)) \subseteq X_1$ (resp. $VARS(hi(v)) \subseteq X_2$). Then Build($\varphi_j$, $i$) can terminate in finite time, $VARS(v) \subseteq X$, the conditions in Definition 1 and Definition 3 are satisfied. Lines 4-8 guarantee the condition in Definition 5. The loop on Lines 10-16 guarantees that $\varphi \mid \neg x_j$ is not equivalent to $\varphi \mid x_j$. By Proposition 2, $lo(v)$ is not identical to $hi(v)$. The call of MK guarantees the condition 1 in Definition 4. And the loop must terminate at some $j$, otherwise the condition on Line 17 will be satisfied, which corresponds to the case that $\varphi$ is equivalent to *true* after Line 9. Then the output of Build($\varphi$, X, $i$) is the ROBDD-$i$ corresponding to $\varphi$. ■

```
procedure Build(φ, i)
     // φ is a Boolean formula over the variables x_1 < … < x_n
1:   if φ is unsatisfiable then v ← ⊥
2:   else
3:   |    Create a new node v
4:   |    for j = 1 to n and | L(v) | < i do
5:   |         if φ implies x_j then L(v) ← L(v) ∪ {x_j}
6:   |         else-if φ implies ¬x_j then L(v) ← L(v) ∪ {¬x_j}
7:   |         endif
8:   |    endfor
9:   |    φ ← φ | L(v)
10:  |    for j = 1 to n do
11:  |         if x_j cannot be omitted in φ then
12:  |              v ← ⟨x_j, Build(φ | ¬x_j, i), Build(φ | x_j, i), L(v)⟩
13:  |              break
14:  |         else φ ← φ | x_j
15:  |         endif
16:  |    endfor
17:  |    if j > n then v ← ⟨L(v)⟩ endif
18:  endif
19:  MK(v)
20:  return v
```

Figure 10. The algorithm Build

It is pointed out that Line 11 in the algorithm Build is not needed any more if $i = 0$ or $\infty$, and we just need to call the algorithm Reduce to turn the result into ROBDD or ROBDD-$\infty$. Particularly, we do not even need to Line 1 and Lines 4-9 if $i = 0$, then Build is equivalent to Algorithm 5 in [20] when the input is in CNF. We can displace the algorithm Build with the algorithm Build-inf in Figure 11 if $i = \infty$. In the algorithm Build-inf, Decide($\varphi$) will return a model of $\varphi$ if it is satisfiable, otherwise the empty set will be returned. It is trivial to prove that it is correct to use the function Get-imps to compute all literals implied by $\varphi$.

```
procedure Build-inf(φ)
      // φ is a Boolean formula over the variables x_1 < … < x_n
1:    function Get-imps(φ, Ω)    // this function also output Ω
2:    |     C ← {l : ∀(M ∈ Ω).(¬l ∉ M)}
3:    |     C′ ← ∅
4:    |     while there exists some literal l such that l ∈ C \ C′ do
5:    |           M ← Decide(φ ∪ {¬l})
6:    |           if M ≠ ∅ then Ω ← Ω ∪ {M}; C ← C ∩ M
7:    |           else C′ ← C′ ∪ {l}
8:    |           endif
9:    |     endwhile
10:   |     return C′
11:   end function Get-Imps
12:   function Build-inf-sub(φ, Ω)
13:   |     Create a new node v
14:   |     L(v) ← Get-Imps(φ, Ω)
15:   |     φ ← φ | L(v)
16:   |     if φ is true then v ← ⟨L(v)⟩
17:   |     else
18:   |           Let x be the least variable in φ
19:   |           Ω_1 ← {M \ ({x} ∪ L(v)} : x ∈ M}
20:   |           Ω_2 ← {M \ ({¬x} ∪ L(v)} : ¬x ∈ M}
21:   |           v ← ⟨x, Build(φ | ¬x, Ω_1), Build(φ | x, Ω_2), L(v)⟩
22:   |     endif
23:   end function Build-inf-sub
24:   M ← Decide(φ)
25:   if M = ∅ then return ⊥
26:   return Reduce(Build-inf-sub(φ, {M}))
```

Figure 11. The algorithm Build-inf

Obviously, by using Get-imps, we can significantly reduce the number of calling the function Decide. In fact, the efficiency of the algorithm is heavily dependent on the calling number and efficiency of Decide. So Get-imps is helpful for improving the efficiency. It is pointed out that we can limit the literals in $C$ to the ones appearing in $\varphi$ on Line 2 if $\varphi$ is in NNF because a NNF formula can only imply the literals appearing it. By some proof analogous to the one of Proposition 10, the algorithm Build-inf is correct:

**Proposition 11**. Given any propositional formula $\varphi$, Build-inf($\varphi$) can terminate in finite time and its output is the corresponding ROBDD-$\infty$.

### 5.2 Some Techniques to Improve the Performance of Build-inf

The algorithm Build-inf in Figure 11 is adapted to any propositional formula. However, knowledge base often is represented as a CNF formula as human can read and write it with ease. Here are some techniques which are potential to improve the performance of Build and Build-inf to compile a CNF formula: (1) aims at reducing the number of calling the function Decide, (2) is used to reduce the time of such calling, and the purpose of (3) is to reduce the number of recursive calling of Build-inf-sub.

(1) Horn lower approximation

As previously pointed out, the efficiency of the algorithm is heavily dependent on the calling number of Decide. Given a CNF formula $\varphi$, a Horn lower approximation is a Horn theory implying $\varphi$, so $\varphi$ implies a literal only if any Horn lower approximation of $\varphi$ implies it. And it is well known that all implied literals of Horn theory can be computed in polytime. We exploit these properties to reduce the number of calling Decide, i.e., the function Get-imps will be displaced by the function Get-imps-CNF in Figure 12. We give an example to show how Get-imps-CNF works.

**Example 2**. Let $\varphi = \{x_1 \vee x_3, \neg x_2 \vee x_3, \neg x_1 \vee \neg x_4, x_3 \vee x_4\}$ and $\Omega = \emptyset$. $C = \{x_1, \neg x_1, x_2, \neg x_2, x_3, \neg x_3, x_4, \neg x_4\}$ after running Line 11. Obviously, $\varphi$ does not imply $x_1$. Assume that Decide($\varphi \cup \{\neg x_1\}$) on Line 14 returns a model $M = \{\neg x_1, \neg x_2, x_3, \neg x_4\}$. Then Horn-app($\varphi$, $M$) generates a Horn lower approximation $\varphi' = \{x_3, \neg x_2 \vee x_3, \neg x_1 \vee \neg x_4\}$. $\varphi'$ only implies $x_3$. After running Line 17, $C = \{x_3\}$. Then we find that $\varphi$ also implies $x_3$. Note that we only need to employ SAT solver twice here, while solver will be called at least 3 up to 5 times in Get-imps (dependent on the models obtained by solver).

In fact, a model $M$ of $\varphi$ is also a Horn lower approximation (each literal in $M$ can be seen as a unit Horn clause). However, the approximation generated by Horn-app($\varphi$, $M$) is obviously greater than $M$, i.e., $M \Rightarrow$ Horn-app($\varphi$, $M$), for example, the Horn approximation in Example 2 have 6 models. The idea of Horn theory approximation in [1] is generating a greatest lower bound (GLB). However, [21] showed that this problem is at least $p^{\mathrm{NP}[O(\log n)]}$-hard. It seems that generating a GLB is not desirable in Build-inf.

```
function Get-imps-CNF(φ, Ω)
    function Horn-app(φ, M)
    |   for any C ∈ φ do
    |       if there exists some x ∈ C ∩ M then C′ ← {x}
    |       else-if there exists some x′ in C then C′ ← {x′}
    |       else C′ ← ∅
    |       endif
    |       C ← {¬x | ¬x ∈ C} ∪ C′
    |   endfor
    |   return φ
    end function Horn-app
    C ← {l : ∀(M ∈ Ω).(Horn-app(φ, M) ⇒ l)}
    C′ ← ∅
    while there exists some literal l such that l ∈ C \ C′ do
        M ← Decide(φ ∪ {¬l})
        if M ≠ ∅ then
            Ω ← Ω ∪ {M}
            C ← C ∩ {l : Horn-app(φ, M) ⇒ l}
        else C′ ← C′ ∪ {l}
        endif
    endwhile
    return C′
end function Get-Imps-CNF
```

Figure 12. The function Get-imps-CNF

(2) Employing high-performance SAT solver

We know that we need to employ SAT solver on Lines 1, 5, 6 and 11 in Build. Despite that Build-inf can avoid many times of such employment, efficiency of SAT solver still significantly affect the efficiency of Build-inf (we can see it in Section 6). Fortunately, there exist many efficient complete SAT solvers so far, such as MiniSAT [22], PrecoSAT [23], CryptoMiniSat [24] and so on. All these modern complete SAT solvers are based on the classic DPLL procedure [25], which employs a systematic search to find a model of the inputting CNF formula. In fact, it has been showed that DPLL procedure is closely related to knowledge compilation. For example, [26] proposed an algorithm to exploit DPLL search to generate Horn GLB; [20] proposed a new method to map different versions of exhaustive DPLL search to different compilation languages as ROBDD, FBDD, and d-DNNF, [12] extended this idea to map another version of systematic search to EPCCL theory.

Research in recent years has greatly improved the efficiency and scalability of systematic search methods. Techniques contributing to this improvement include two-literal watch scheme for fast BCP, clause learning by conflict analysis, dependency directed backtracking, new variable ordering heuristics, timely restarts, and so on [20, 22-24]. We emphasize that that clause learning by conflict analysis is very useful here as we always perform DPLL search on the same CNF formula (i.e. knowledge base) only under different initial partial assignments[1], which leads to that learnt clauses can be inherited by other DPLL searches.

(3) CNF caching

In order to save the compiling time, the function Build-inf-sub will be trying to not compiling the same CNF formula twice, we exploit the CNF caching scheme introduced in [27] to do this. Assume the knowledge base needed to be compiled is $\varphi_0 = C_1 \wedge \ldots \wedge C_n$. We know that each input $\varphi$ of Build-inf-sub comes from conditioning $\varphi_0$ on a partial assignment $A$. Given any two inputs $\varphi = \varphi_0 \mid A$ and $\varphi' = \varphi_0 \mid A'$, $\varphi \Leftrightarrow \varphi'$ if the following conditions hold: the variables appearing in $A$ and $A'$ are the same; let $bv$ be a vector with $n$ bit, $bv(i) = 1$ if $C_i$ shares some literal with $A$, otherwise $bv(i) = 0$, $bv'$ is generated from $A'$ in the same way, $bv$ equals to $bv'$. For example, let $\varphi = (x_1 \vee \neg x_2) \wedge (\neg x_1 \vee x_2) \wedge (x_3 \vee \neg x_4)$, we know that $\varphi \mid x_1 \wedge x_2$ is equivalent to $\varphi \mid \neg x_1 \wedge \neg x_2$ by this scheme.

### 5.3 Transforming ROBDD-∞ into ROBDD

As pointed out previously, the algorithm Build can be turned into Algorithm 5 in [20] when the input is in CNF. However, the experimental results in [20] show that the efficiency of Algorithm 5 is needed to be further improved. In this subsection, we present an alternative — we show that any ROBDD-∞ can be transformed into the equivalent ROBDD using the algorithm Inf2ROBDD in Figure 13. And we will see that this approach outperforms Algorithm 5 in [20] in the next section. Note that on Line 10, we just need to displace the node $\langle \emptyset \rangle$ in the ROBDD whose root is $u_2$ with $u_1$ to get the conjunct since any variable in $VARS(u_2)$ is less than the one in $VARS(u_1)$.

**Proposition 12**. Given any ROBDD-∞ with the root $v$, Inf2ROBDD($v$) can terminate in finite time and its output is the corresponding ROBDD.

**Proof**. It is trivial to prove by induction that the output of Inf2R-sub($v$, $T$) is the ROBDD corresponding to $\vartheta(v) \wedge T$, where $VARS(v)$ and $T$ do not share any variable with each other. Then this proposition holds. ■

[1] Both Build-inf and DPLL procedure are implemented iteratively.

```
procedure Inf2ROBDD(v)
    function Inf2R-sub(v, T)
        if G6(v, T) ≠ empty then return G6(v, T) endif
        if v is a True node then
            Let u be the root of the ROBDD corresponding to T ∪ L(v)
        else
            T1 ← {l ∈ T ∪ L(v) : var(v) is less than the variable of l}
            T2 ← T ∪ L(v) \ T1
            u1 ← ⟨var(v), Inf2R-sub(lo(v), T′) Inf2R-sub(hi(v), T′), ∅⟩
            Let u2 be the root of the ROBDD corresponding to T2
            Let u be the root of the conjunct of u1 and u2
        endif
        G6(v, C) ← u
        return u
    end function Inf2R-sub
    if v = ⊥ then return ⊥
    return Reduce(Inf2R-sub(v, ∅))
```

Figure 13. The function Get-imps-CNF

## 6 Preliminary Experimental Results

In this section, we report some experimental results about our ROBDD-*L* package BDDjLu. In BDDjLu, any CNF formula can be compiled into ROBDD-*i* ($0 \le i < \infty$) by the algorithm Build or ROBDD-∞ by the algorithm Build-inf, and all the operations supported by ROBDD-∞ in polytime are included. BDDjLu also contains the algorithms Inf2ROBDD and Inf2FBDD. As pointed out previously, in Build and Build-inf, SAT solver will be employed to decide whether a CNF formula is satisfiable or not, get the implied literals in a CNF formula, decide whether the variables can be omitted in CNF formula. So we also implement a DPLL-based SAT solver based on the one used in [13, 28], in which we exploit some techniques including two-literal watch scheme for fast BCP, clause learning by conflict analysis, dependency directed backtracking, variable ordering heuristics VSIDS.

We compare BDDjLu against a d-DNNF compiler called c2d[2] [26]. ROBDD-∞, FBDD and ROBDD are generated by Build-inf, Inf2FBDD and Inf2ROBDD, respectively. It seems that a ROBDD compiler and a FBDD compiler were implemented by Huang and Darwiche in [20]. But we cannot compare BDDjLu with them. The directives -reduce and -dt-method 4 (i.e., min-fill heuristic for constructing d-trees) are used in c2d. -reduce is helpful to generate a smaller d-DNNF and -dt-method 4 seems to work best on a broad set of benchmarks[3]. In our experiments, the variables are denoted by their indices (i.e. some natural numbers) and $x_j < x_k$ iff $j < k$. All experiments are conducted on a computer with a 2.79 GHz CPU and 512 MB of RAM. The timeout for each problem is set to 1000 seconds.

We test BDDjLu and c2d on some benchmarks from SATLIB[4] and the experimental results are showed in Table 3, where problem types (#vars, #cls and #models indicate the number of variables, clauses and models of corresponding instance, respectively), compilation output sizes of d-DNNF, ROBDD-∞, FBDD and ROBDD (#nodes and #edges indicate the number of nodes and edges, respectively), individual compiling time[5] (- indicates timeout or memory overflow) and the time cost of employing SAT solver in Build-inf (sat indicates it) are reported.

[2] http://reasoning.cs.ucla.edu/c2d/

[3] Knot Pipatsrisawat, a past member in the Automated Reasoning group at UCLA, said these in the private communication between us.

[4] http://people.cs.ubc.ca/~hoos/SATLIB/benchm.html

[5] For FBDD (resp. ROBDD), it is the sum of the time cost of Build-inf and Inf2FBDD (resp. Inf2ROBDD).

Table 3. Comparison between d-DNNF, ROBDD-∞, FBDD and ROBDD

| CNF instance | #vars | #cls | #models | d-DNNF | | | ROBDD-∞ | | | | FBDD | | | ROBDD | | |
|---|---|---|---|---|---|---|---|---|---|---|---|---|---|---|---|---|
| | | | | time/s | #nodes | #edges | sat/s | time/s | #nodes | #edges | time/s | #nodes | #edges | time/s | #nodes | #edges |
| CBS_k3_n100_m403_b10_0 | 100 | 403 | 214200 | 0.23 | 509 | 2636 | 0.08 | 0.09 | 421 | 786 | 0.11 | 1288 | 2452 | 0.12 | 12876 | 25748 |
| CBS_k3_n100_m403_b50_1 | 100 | 403 | 11408 | 0.16 | 258 | 1072 | 0.03 | 0.05 | 191 | 324 | 0.07 | 689 | 1374 | 0.08 | 3580 | 7156 |
| CBS_k3_n100_m403_b90_2 | 100 | 403 | 144 | 0.19 | 113 | 281 | 0.05 | 0.06 | 10 | 16 | 0.06 | 102 | 200 | 0.06 | 416 | 828 |
| flat200-1 | 600 | 2237 | 5.38e11 | 34.78 | 60343 | 262467 | - | - | - | - | - | - | - | - | - | - |
| flat200-2 | 600 | 2237 | 1.37e10 | 107.42 | 2061 | 16370 | 12.60 | 13.91 | 11765 | 23522 | 13.94 | 31716 | 63428 | 20.35 | 873516 | 1747028 |
| flat200-3 | 600 | 2237 | 1.52e10 | 87.50 | 3124 | 14364 | 8.92 | 13.34 | 43921 | 87834 | 13.43 | 111593 | 223182 | 22.7 | 944373 | 1888742 |
| hole8 | 72 | 397 | 0 | 3.27 | 1 | 0 | 0.08 | 0.08 | 1 | 0 | 0.08 | 1 | 0 | 0.08 | 1 | 0 |
| hole9 | 90 | 415 | 0 | 82.17 | 1 | 0 | 0.95 | 0.97 | 1 | 0 | 0.97 | 1 | 0 | 0.97 | 1 | 0 |
| hole10 | 110 | 561 | 0 | - | - | - | 10.06 | 10.08 | 1 | 0 | 10.08 | 1 | 0 | 10.08 | 1 | 0 |
| par8-2-c | 68 | 270 | 1 | 0.09 | 69 | 68 | 0.00 | 0.00 | 1 | 0 | 0.02 | 70 | 136 | 0.03 | 70 | 136 |
| par16-2-c | 349 | 1392 | 1 | 3.75 | 350 | 349 | 3.63 | 3.67 | 1 | 0 | 3.72 | 351 | 698 | 3.76 | 351 | 698 |
| par32-2-c | 1303 | 5206 | - | - | - | - | - | - | - | - | - | - | | - | - | |
| uf200-01 | 200 | 860 | 112896 | 148.72 | 245 | 739 | 5.61 | 5.66 | 19 | 28 | 5.71 | 211 | 418 | 5.8 | 1833 | 3662 |
| uf200-02 | 200 | 860 | 1555776 | 32.41 | 299 | 396 | 4.84 | 4.92 | 176 | 338 | 4.95 | 460 | 916 | 5.05 | 16441 | 32878 |
| uf200-03 | 200 | 860 | 8.04e8 | 132.56 | 22941 | 72249 | 50.80 | 61.70 | 46695 | 92778 | 61.83 | 134814 | 269624 | 124.51 | 2649680 | 5299356 |
| sat-grid-pbl-0015 | 110 | 191 | 3.01e54 | 0.36 | 3368 | 6791 | 4.56 | 4.78 | 538 | 1072 | 4.8 | 672 | 1340 | 4.8 | 604 | 1204 |
| sat-grid-pbl-0020 | 420 | 781 | 5.06e95 | 286.81 | 640132 | 1445683 | - | - | - | - | - | - | - | - | - | - |
| sat-grid-pbl-0025 | 650 | 1226 | - | - | - | - | - | - | - | - | - | - | - | - | - | - |

The results in Table 3 show that the compilation quality of BDDjLu for ROBDD-∞ is higher than c2d for CBS_k3_n100_m403_*, par8-2-c, par15-2-c, uf200-01, uf200-02 and sat-grid-pbl-0015. And c2d is more high-quality for flat200-2, flat200-3 and uf200-03. c2d cannot compile hole10, while BDDjLu cannot compile flat200-1 and sat-grid-pbl-0020 [6]. Neither c2d nor BDDjLu can compile par32-2-c and sat-grid-pbl-0025. Compared with the experimental results in [20], the FBDD generated by Inf2FBDD is obviously smaller for flat200-* and uf200-*. For all instances that can be compiled into ROBDD-∞, the size of ROBDD-∞ is obviously small than FBDD, while the size of FBDD is obviously smaller than ROBDD except sat-grid-pbl-0015. These validate the assertion in Section 3. Turning to running time, BDDjLu for ROBDD-∞ is faster than c2d except flat200-1 and sat-grid-pbl-* [7]. flat200-2, flat200-3 and uf200-* can be compiled by transforming the corresponding ROBDD-∞ into ROBDD now, while the ROBDD compiler reported in [20] cannot compile them in 900s with a 2.4 GHz CPU and 4 GB RAM.

Overall, the ROBDD-∞ is obviously smaller than the ROBDD for each instance. The algorithm Build-inf in BDDjLu is more high-quality than c2d for the benchmark corresponding to small ROBDD-∞, and it is faster for most of problems when c2d uses min-fill heuristic to construct d-trees. Compared with the compilers reported in [20], it seems that Inf2FBDD and Inf2ROBDD are two good alternatives to compile CNF formulas into FBDDs and ROBDDs, respectively. As the time cost of employing SAT solver in Build-inf plays a very important role in the total running time of generating ROBDD-∞, we shall straight embed some highly efficient modern SAT solver into BDDjLu. In addition, we shall design good variables order rather than current simple order in BDDjLu to generate more space-efficient compilation results in future.

Finally, we note that, in comparison with d-DNNF, FBDD and ROBDD, we need a little additional processing on the implied literals of ROBDD-∞ in the querying. Fortunately, on the one hand, some operations (e.g., model counting, computing the minimum cardinality) only need some fragment information about implied literals, therefore, we can visit any node of ROBDD-∞ in constant time; on the other hand, we can reduce the number of processing for other operations with the use of implied literals table.

## 7 Conclusions

In this paper, we introduce a new compilation approach ROBDD-*L* by associating some implied literals in each node of ROBDD. Then an interesting kind of subsets of ROBDD-*L* is discussed: given a number *i*, we call the corresponding subset ROBDD-*i*, which requires that all of its nodes should be precisely associated by *i* implied literals. In particular, the ROBDD-0 whose nodes have no implied literal is isomorphic to ROBDD; the ROBDD-∞ requires that every node should be associated by the implied literals as many as possible. Given a number *i* and a Boolean formula, we show that there is exactly one ROBDD-*i* representing it over a specific variables order.

Furthermore, we show that every sentence of OBDD-*L* can be transformed into an equivalent sentence of ROBDD-∞ in polytime. This means that ROBDD-∞ is the most succinct subset of OBDD-*L*. Particularly, ROBDD-∞ is strictly more succinct than ROBDD. Compared with FBDD and d-DNNF, ROBDD-∞ is strictly less succinct. And we propose the algorithm Inf2FBDD which can transform any ROBDD-∞ into FBDD. In order to evaluate the inferential power of this interesting subset, we compare it with ROBDD, FBDD and d-DNNF by the operations that can be supported in polytime with respect to the knowledge compilation map. For the queries, ROBDD satisfies **SE** which ROBDD-∞ does not satisfy, while ROBDD-∞ satisfies **EQ** which FBDD and d-DNNF does not satisfy. For the transformations, ROBDD satisfies **CD**, **SFO,** ∧**BC,** ∨**BC** and ¬**C**, both ROBDD-∞ and d-DNNF only satisfy **CD**, and FBDD satisfies **CD** and ¬**C**.

Finally, we propose the compilation algorithm Build which can compile any Boolean formula into a ROBDD-*i* for any *i*. Based on it, we propose the ROBDD-∞ compilation algorithm Build-inf and discuss three optimization techniques. In addition, we show that every ROBDD-∞ can be transformed into ROBDD by proposing the algorithm Inf2ROBDD. Combining Build, Build-inf, Inf2FBDD, Inf2ROBDD and all the operations supported by ROBDD-*L* in polytime, we devise the ROBDD-*L* package BDDjLu and test it on some benchmarks from SATLIB. Preliminary experimental results show that: for the same instance, ROBDD-∞ is obviously smaller than ROBDD; Build-inf in BDDjLu is more high-quality than c2d for the benchmarks corresponding to small compilation results, and it is faster for most of problems when c2d uses min-fill heuristic to construct d-trees; it seems that it is better to transform the ROBDD-∞ into FBDD and ROBDD using Inf2FBDD and Inf2ROBDD. And both efficiency and compilation quality of BDDjLu have potential to be improved by embedding a modern SAT solver and devising a better variables order.

**Acknowledgements** We thank Professor Bart Selman for the suggestions about this paper. We also thank Dr. Knot Pipatsrisawat for the useful information about c2d provided by him. Finally, we thank the anonymous referees for their review of this paper. Our work is supported by the NNSF of China grant 60873149 and 60973088.

[6] For sat-grid-pbl-0020, BDDjLu can generate a ROBDD-∞ with 2358 nodes and 4712 edges in 62.56s when the heuristic value does not decay in our SAT solver, and the resulting ROBDD-∞ can be transformed into FBDD (resp. ROBDD) with 3092 (resp. 2512) nodes and 6180 (resp. 5020) edges in 0.02s (resp. 0.03s).

[7] It seems that , by selecting a different method of generating d-tree, c2d compiles flat200-* and uf200-* into d-DNNF faster in [?].